\documentclass[conference]{IEEEtran}
\usepackage{microtype}
\usepackage{graphicx}
\usepackage{booktabs} % for professional tables
\usepackage{cite}      % Pour gérer les citations numériques
\usepackage{tabularx}
\usepackage{multirow}

\usepackage{amsmath,amssymb,amsfonts}
\usepackage{textcomp}
\usepackage{xcolor}
\usepackage{amsthm}
\usepackage{hyperref}

\def\BibTeX{{\rm B\kern-.05em{\sc i\kern-.025em b}\kern-.08em
    T\kern-.1667em\lower.7ex\hbox{E}\kern-.125emX}}

\newtheorem{definition}{Definition}

\title{
Risk Is Not the Target: A Monotonic Framework for Evaluating Wildfire Operational Risk Signals
}

\author{
\IEEEauthorblockN{Nicolas Caron, Christophe Guyeux, Hassan Noura, Maxime Coulmeau}
\IEEEauthorblockA{
Universit\'e Marie et Louis Pasteur, CNRS, institut FEMTO-ST (UMR 6174) \\
F-90000 Belfort, France \\
\{nicolas.caron, christophe.guyeux, hassan.noura, maxime.coulmeau\}@umlp.fr
}
\and
\IEEEauthorblockN{Benjamin Aynes}
\IEEEauthorblockA{
SAD Marketing \\
b.aynes@sad-marketing.com
}
}

\begin{document}
\maketitle

\begin{abstract}
	Evaluating wildfire risk systems using standard machine-learning metrics such as F1-score or IoU is fundamentally flawed: these metrics assess event prediction accuracy, not the operational coherence of a continuous risk signal. This work proposes a novel monotonic evaluation framework that measures whether increases in a predicted risk score consistently correspond to increases in observed operational load, such as number of fires, intervention time, and deployed resources. Moreover, we compare three structurally different approaches on the French Alpes-Maritimes department: the expert-based DFE index, GRU-based predictive models, and FARS, a hybrid multi-agent system combining predictive AI with LLM-based reasoning. Experimental results reveal that the DFE, despite poor classification metrics, exhibits the most balanced monotonic behavior across the full risk scale. GRU models achieve strong local monotonicity but fail to produce well-distributed risk levels. FARS inherits and reveals the structural limitations of upstream signals rather than correcting them. The central finding is a paradigm shift: a good risk model does not predict fires accurately, but one whose ordinal scale meaningfully explains operational dynamics, as proved in this paper.
    Code of the monotonic framework is available on~\href{https://github.com/NicolasCaronPro/Risk-Is-Not-the-Target-A-Monotonic-Framework-for-Evaluating-Wildfire-Operational-Risk-Signals/tree/main}{github}.
\end{abstract}

\section{Introduction}
Predicting forest fire risk in France has become a strategic priority for improving prevention, preparedness, and rapid operational response. According to the European Forest Fire Information System (\href{https://joint-research-centre.ec.europa.eu/jrc-news-and-updates/eu-2022-wildfire-season-was-second-worst-record-2023-05-02_en}{JRC/EFFIS}), the year 2022 was the \emph{second-worst} wildfire year since 2000. Recurrent heatwaves, prolonged droughts, and cumulative fuel drying have led to repeated extreme fire seasons in countries such as Spain, Portugal, Greece, and France.

Climate change is a major driver of this escalation, as it lengthens the fire season, reduces fuel moisture, and increases the likelihood of extreme fire-weather conditions. These mechanisms are well documented in national scientific assessments, which show that climate change is amplifying both the frequency and severity of forest fires and extending risk to new regions (\href{https://meteofrance.com/le-changement-climatique/quel-climat-futur/changement-climatique-quel-impact-sur-les-feux-de-foret}{Météo-France}; \href{https://www.inrae.fr/actualites/dereglement-climatique-attise-risques-feux-forets}{INRAE}).

Several operational products already support authorities and the public. In particular, \emph{Météo des Forêts}, developed by Météo-France, provides daily wildfire danger maps at the departmental level for the next two days using a four-level danger scale (\href{https://meteofrance.com/meteo-des-forets}{Météo des Forêts}). While such tools are essential for awareness and preparedness, they remain limited in their ability to directly support operational decision-making for firefighting agencies, as they are primarily based on meteorological conditions.

AI has been widely used to overcome the limitations of traditional statistical risk indices by incorporating a broader range of input factors, such as land cover, socio-economic factors, and, more recently, with the emergence of LLMs, textual data. %First, AI-based tools have been shown to outperform traditional statistical methods in several contexts. In addition, the recent widespread adoption of large language models (LLMs) enables end users, such as firefighters and government agencies, to obtain rapid, textual, and understandable responses, thereby facilitating timely and informed decision-making.

\subsection{State of the Art}

One of the most widely used operational indices is the Fire Weather Index (FWI)~\cite{vanwagner1987fwi, Varela, Podschwit2022, https://doi.org/10.1002/eap.2316}, employed to classify danger levels for readiness and resource allocation. However, it relies solely on meteorological factors and ignores socio-economic influences.

AI-based approaches have been explored for wildfire risk mapping~\cite{WildfireDangerPrediction}, daily danger prediction~\cite{karasante2023seasfiremultivariateearthdatacube, michail2025firecastnetearthasagraphseasonalprediction}, and fire spread simulation. A primary limitation is that these studies do not account for operational load: available resources, deployed vehicles, and intervention times, variables critical for daily preparedness at fire stations.

Additionally, these studies overlook a fundamental distinction: events are discrete, whereas risk is continuous. Indices like the FWI do not predict whether a fire will occur, but indicate potential severity should a fire ignite. Direct comparisons between AI event prediction and statistical risk indices~\cite{WildfireDangerPrediction} thus disadvantage the latter. High classification performance does not demonstrate that the AI-derived measure is operationally meaningful for anticipating resource demand.

LLMs have only recently emerged in wildfire management. WildfireGPT~\cite{xie2024wildfiregpttailoredlargelanguage} synthesizes historical data and scientific literature for contextual guidance. Dolant et al.~\cite{dolant2025agenticllmframeworkadaptive} introduce an agentic LLM framework for decision support, while Chen et al.~\cite{chen2025empoweringllmagentsgeospatial} demonstrate geospatial grounding for resource estimation during ongoing fires. However, none produce quantitative, time-resolved forecasts of operational risk. The conceptualization of multidimensional risk has been previously articulated in~\cite{caron2026proofconceptmultitargetwildfire}, proposing a continuous risk prediction system combining predictive and generative AI. Although introduced conceptually, no study to date has explored LLMs producing an operational wildfire risk assessment.

\subsection{The Evaluation Gap}

A critical issue in wildfire risk prediction is the mismatch between risk signals and evaluation metrics. Standard metrics (F1-score, IoU, AUC) assess event prediction, but are structurally inadequate for risk signals. First, risk is a continuous latent variable, whereas fire events are discrete and stochastic. A high-quality risk signal indicates expected operational pressure, not whether a specific fire will occur. Second, metrics like F1 and IoU reward models that fit discrete targets regardless of whether predictions form a meaningful ordinal structure. A model may achieve high classification performance while producing an operationally meaningless risk scale. A fire modeling framework aligned with classification metrics such as the F1 score is more consistent with spatial fire prediction, in which each pixel is assigned a probability of occurrence such as ~\cite{10633472}. However, in such a setting, the calibration of these probabilities must be carefully assessed. This objective differs from the one considered in the present article.

This evaluation gap has profound implications: models optimized for classification may fail to support operational decision-making. The need for monotonic coherence is well established in the literature through isotonic regression~\cite{barlow1972isotonic} and probability calibration~\cite{platt1999probabilistic, zadrozny2002transforming}. The present study adapts these principles by proposing a framework that assesses risk signals based on their monotonic alignment with observed operational load, rather than their ability to predict discrete events.

\subsection{Contributions}
AI has demonstrated clear value in wildfire management, but constructing a continuous measure of operational risk, a value explaining increases in operational load rather than predicting event occurrence, remains an unresolved challenge.

This article proposes a complete methodology for designing an operational wildfire risk system and an evaluation framework capable of revealing the real limitations of current approaches. We introduce a monotonic evaluation framework that assesses whether risk score increases consistently correspond to operational load increases.

Using this framework, we compare structurally different risk systems: a predictive AI model, an expert-based statistical index (the DFE), and a hybrid predictive--generative AI architecture (FARS, Fire Agent Risk System, previously articulated in the literature~\cite{caron2026proofconceptmultitargetwildfire}). The objective is to reveal their structural limitations when evaluated through a common monotonic lens.

\textbf{Central thesis:} A good risk model does not predict fires accurately, but one whose ordinal scale meaningfully explains operational dynamics.

\subsection{Organization}
The remainder of this article is organized as follows. Section~\ref{sec:dataset} describes the dataset and the region of study. Section~\ref{sec:mono} details the proposed monotonic evaluation framework. Section~\ref{sec:candidates} introduces the risk modeling approaches, including expert, statistical, and AI-based methods. Section~\ref{sec:eval} presents the experimental setup and results. Finally, Section~\ref{sec:concl} concludes the study and discusses future perspectives.

\section{Dataset} \label{sec:dataset}

This section concisely describes the dataset used in this study.

\subsection{Region of Study}

This work focuses on the French department of Alpes-Maritimes (06), a region marked by strong contrasts between densely populated Mediterranean coastal cities and sparsely inhabited, forested inland and mountainous areas. This diversity results in spatially heterogeneous wildfire risk profiles, shaped by both environmental and human factors.

\begin{figure}[ht]
	\centering
	\includegraphics[width=\linewidth]{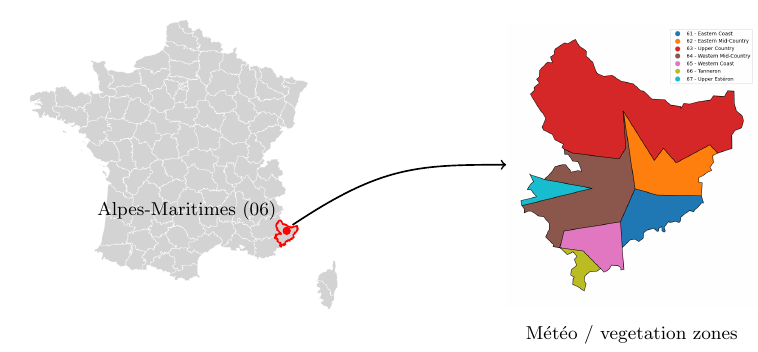}
	\caption{Position of the Alpes-Maritimes in France and its segmentation into meteorological zones (Figure adapted from publicly available material described in~\cite{caron2026proofconceptmultitargetwildfire}).}
	\label{fig:region}
\end{figure}

For operational modeling, the department is partitioned into six meteorological zones, as defined by Météo-France (see Figure~\ref{fig:region}). Each meteorological zone in the Alpes-Maritimes corresponds to distinct vegetation profiles, ranging from sparse coastal shrublands and Aleppo pine stands in urbanized areas to dense broadleaf or Scots pine forests and alpine larch in inland and mountainous regions.

\subsection{Targets and Features}

\subsubsection{Predictive Targets}
Wildfire risk is modeled using three complementary targets, each capturing a distinct operational aspect:
\begin{itemize}
	\item \textbf{Number of Fires:} The daily count of wildfire events per zone (data from 2017 to 2023).
	\item \textbf{Intervention Time:} The total duration (in minutes) of firefighting operations per day and zone (data from 2017 to 2023).
	\item \textbf{Resources:} The number of firefighting units deployed (data from 2017 to 2023).
\end{itemize}

Wildfire risk prediction is multidimensional because it combines several complementary and weakly correlated dimensions, ignition activity, intervention complexity, and resource mobilization, each reflecting distinct environmental, human, and organizational processes. With these three operational variables, we obtain a range of risk levels, each reflecting different forms of operational load and constraints on firefighting services.

Importantly, operational variables such as Number of Fires, Intervention Time, and Resources exist in two forms throughout this work: (i) a continuous observed value (counts, minutes, deployed units), and (ii) an ordinal representation obtained by clustering, used only as the supervised target for the predictive models. The monotonic evaluation framework always uses the continuous observed values as outcomes and evaluates whether the predicted ordinal risk levels behave consistently with these continuous outcomes.

\subsubsection{Feature Set}

We used a feature set described in~\cite{caron2025localizedforestriskprediction}. The data are grouped into four main categories: meteorological (e.g., FWI indices, temperature, humidity, wind), topographic (land cover, satellite-derived indices), calendar-related (day of the year, weekends, holidays), socio-economic (population density), and historical (last observed risk levels). The features are initially collected at a $2\,\mathrm{km} \times 2\,\mathrm{km}$ spatial resolution. After encoding the categorical variables, they are aggregated using the mean, maximum, and minimum, resulting in a single feature set for each zone and date.

\subsubsection{Dataset Split}

For each target (Number of Fires, Intervention Time, and Resources), a strict year-based data split was used to avoid temporal data leakage. The operational targets were trained on data from 2017 to 2020 and 2022.

The year 2021 was used for validation and hyperparameter tuning, and 2023 was kept as a fully independent test set. The year 2022 was included in the training data because its exceptional wildfire severity provided valuable extreme-event examples for model learning.

All preprocessing steps and target transformations were learned only on the training data and applied unchanged to the validation and test sets.

\section{Monotonic Evaluation of Risk Signals} \label{sec:mono}
This section presents the risk system evaluation framework proposed in this study.

\subsection{Overview of the Risk Evaluation}
In most domains, risk analysis is fundamentally framed as a probability problem~\cite{ustun2019learningoptimizedriskscores}. Whether in medicine, finance, insurance, or reliability engineering, a risk score is typically interpreted as an estimate of the probability that a specific adverse event will occur.

Within this probabilistic paradigm, the goal of a risk score is to answer a binary-oriented question: will the event happen or not? Consequently, metrics like AUC, Brier score, or calibration plots are used to compare predicted probabilities with observed outcomes. At the same time, several studies have pointed out that classification metrics such as the F1-score can be misleading when used to evaluate risk models, because they emphasize event prediction performance rather than the quality of the risk signal itself~\cite{vancalster2024performanceevaluationpredictiveai}.

Wildfire analysis, and even more so the notion of operational load considered here, belongs to a fundamentally different problem space. The probabilistic view of risk implicitly assumes two key principles: (i) the objective is to predict the occurrence of an event, not the magnitude or number of events, and (ii) the event is well defined, and we know precisely when and why it occurs, so it can be used as a reliable ground truth. Neither of these assumptions holds when analyzing operational wildfire management.

First, the goal is not to determine whether a fire will occur, but rather to anticipate how much operational activity will take place: how many vehicles will be deployed, how much intervention time will be required, how many incidents will demand attention. This is a loaded question, not about the occurrence of a single event.

Second, the occurrence of a wildfire is inherently stochastic. We do not know how many fires could have occurred under the same conditions, nor can we fully attribute their ignition to observable variables. What we observe is already the result of human response, environmental randomness, and preventive actions. Therefore, the observed event cannot serve as a reliable ``ground truth'' for the latent level of danger.

Meteorological fire indices such as the FWI are continuous: they do not attempt to predict fires, but rather to express a degree of environmental severity. They indicate how critical the situation is, not whether an event will happen. Most of the literature evaluates the FWI in a predictive or comparative setting. A first line of work studies its statistical association with wildfire activity, for instance, through ignition probability or burned-area relationships, often using regression-based frameworks~\cite{Beccari2016}. A second line evaluates the FWI in an event-oriented setting, asking whether high FWI values coincide with actual fire events or high-danger days, and therefore relies on discrimination metrics such as probability of detection and related warning-skill measures~\cite{DiGiuseppe2020}. A third line of work emphasizes that the interpretation of a given FWI value is region-dependent, showing that fixed thresholds may not transfer reliably across contrasted climatic and ecological contexts, which motivates regional calibration and distributional validation~\cite{Podschwit2022}.

This article shifts the evaluation target from \emph{event prediction} to \emph{risk-scale}. The objective is no longer to assess how well a model predicts events, but to determine whether a risk scoring system, naturally ordinal in nature, is capable of translating increases in operational pressure. In other words, the question is not: does the score predict fires? But rather: does an increase in the score consistently correspond to an increase in operational load?

\subsection{Proposal}
Evaluating wildfire risk is inherently challenging because the quantity to be measured, a latent, continuous risk signal, does not share the same nature as the observable operational outcomes used for validation. While most studies assess predictive performance by comparing predicted and observed values, this approach becomes limited when the targets themselves are highly stochastic and discrete, such as fire occurrences or intervention activities, in contrast to more deterministic meteorological indicators like DFE (see Definition~\ref{def:dfe}). Throughout this study, risk scores are represented on a five-class ordinal scale (0–4) to align with the DFE framework.

This structural mismatch motivates the use of a monotonic consistency framework instead of standard machine-learning metrics: the objective is not to predict interventions directly, but to verify that the proposed risk score behaves as a true risk signal; when it increases, the observed operational load should increase consistently. Naïve evaluations based on average intervention values per predicted class fail because they confound the effect of the score with spatio-temporal biases and depend heavily on the distribution of predicted classes, thereby masking the true quality of the risk signal.

An evaluation framework for risk scoring systems, based on a
\emph{monotonic consistency analysis}, is introduced.
The relationship between the predicted risk score and the target variable is estimated
using a flexible spline regression with fixed effects (date and zone), as shown in
Equation~\ref{eq:spline_fe}, where $Z$ and $D$ denote zone and date fixed effects, $Y$ denotes the \emph{continuous observed} operational outcome (e.g., number of fires, intervention minutes, deployed resources), while the argument $S$ denotes a \emph{predicted ordinal risk level}. This specification allows the recovery of a smooth response function along the risk
scale while controlling for spatial and temporal heterogeneity.
A spline formulation is preferred to a linear model,
as it enables the detection of local non-linearities and potential decreases in the
response function that may be averaged out by a single global slope.

\begin{align}
	Y \;=\; f\!\left(S\right) \;+\; Z{} \;+\; D{} \;+\; \varepsilon,
	\label{eq:spline_fe}
	\\
	f(\cdot)=\mathrm{B\text{-}spline}(\cdot),
\end{align}

Because $Z$ and $D$ are included as fixed effects, the evaluation is performed \emph{within} the observed spatio-temporal support: it isolates the association between the risk score and the operational outcome after absorbing time-specific and zone-specific variability. This specification avoids comparing outcomes across different dates or zones (e.g., $D_1$ vs.\ $D_2$ or $Z_1$ vs.\ $Z_2$) and focuses the analysis on how the score behaves conditional on the fixed effects.

Model-based mean outcomes are computed at discrete risk
levels by averaging predictions over the observed spatio-temporal support, as reported
in Equation~\ref{eq:mu_s}, with $N$ the number of samples.
These quantities summarize the expected outcome associated with each risk level.

\begin{equation}
	\mu(s) \;=\; \frac{1}{N}\sum_{i=1}^{N}(\widehat{f}(s)\;+\;\widehat{Z}_{z(i)}\;+\;\widehat{D}_{d(i)}).
	\label{eq:mu_s}
\end{equation}

Monotonic behaviour is assessed through \emph{risk transitions} indexed by $k$, where each $k$
groups a set of sub-transitions $P_k=\{(a,b)\}$ (e.g., $P_1=\{(0,1),(1,2),(2,3),(3,4)\}$,
$P_2=\{(0,2),(1,3),(2,4)\}$, etc.). For each sub-transition $(a,b)\in P_k$, the estimated effect is $\Delta_{a\rightarrow b} = \mu(b)-\mu(a)$.
\label{eq:delta_ab}

For each transition order $k$, the summary statistics are defined in Table~\ref{tab:metrics_def}, where $\mathrm{MIN}$, $\mathrm{MED}$, $\mathrm{VIOL}$, and $\mathrm{NEG}$ represent the worst-case behavior, median risk gain, violation frequency, and violation magnitude respectively. They are computed over the set of evaluable sub-transitions.

\begin{table*}[ht]
\centering
\caption{Definition of the monotonic metrics for a transition order $k$.}
\label{tab:metrics_def}
\begin{tabularx}{\textwidth}{lXl}
\toprule
Metric & Value & Definition \\
\midrule
$\mathrm{MED}_k$ & $\mathrm{median}\Big\{\, \Delta_{a\rightarrow b} \;\big|\; (a,b)\in P_k^\star \Big\}$ & Typical monotonic gain \\
$\mathrm{MIN}_k$ & $\min_{(a,b)\in P_k^\star}\Delta_{a\rightarrow b}$ & Worst-case transition \\
$\mathrm{VIOL}_k$ & $\frac{1}{|P_k^\star|}\sum_{(a,b)\in P_k^\star}\mathbb{I}\!\left[\Delta_{a\rightarrow b}<0\right]$ & Frequency of violations \\
$\mathrm{NEG}_k$ & $\frac{1}{|P_k^\star|}\sum_{(a,b)\in P_k^\star}\max\!\left(0,-\Delta_{a\rightarrow b}\right)$ & Magnitude of inverted transitions \\
\bottomrule
\end{tabularx}
\end{table*}

These components are combined into a single transition-level score, according to
Equation~\ref{eq:score_k}. A positive score indicates that the risk signal exhibits a globally consistent monotonic relationship with the observed operational load for
this transition order, whereas a negative score reveals that violations or incoherent gains dominate, indicating that the score fails to behave as a reliable operational risk indicator.

\begin{equation}
\scriptsize
\mathrm{SCORE}_k
=
\frac{\mathrm{MED}_k + \mathrm{MIN}_k}{2}
-
\mathrm{NEG}_k\left(1+\mathrm{VIOL}_k\right).
\label{eq:score_k}
\end{equation}

The proposed score summarizes the set of transition effects
\(\{\Delta_{a\to b}\}_{(a,\;b)\in P_k^\star}\) through four complementary quantities.
\(MED_k\) captures the typical monotonic gain associated with an increase in the risk level and is preferred to the mean because transition amplitudes are not directly comparable across the ordinal scale.
\(MIN_k\) controls the worst supported transition and prevents a globally positive trend from masking a locally inconsistent ordering.
In contrast, \(NEG_k\) measures the average magnitude of inverted transitions, while \(VIOL_k\) measures how frequently such inversions occur.
The penalty term \(NEG_k(1+VIOL_k)\) therefore increases when monotonic failures are both severe and recurrent.
As a result, the score rewards signals that are globally increasing while explicitly penalizing local contradictions, which is consistent with evaluating the structural coherence of an ordinal operational risk scale.

The empirical support available to evaluate a transition is quantified by the coverage $\mathrm{coverage}_k = |P_k^\star|$, which corresponds to the number of sub-transitions that can be effectively evaluated for a given configuration. 

%To prevent some transition orders from numerically dominating others, two aggregated scores are defined, as shown in Equation~\ref{eq:score_low_high}.

%\begin{align}
%\mathrm{SCORE}_{\text{LOW}} \;=\; \sum_{k\in(1,2)}\mathrm{SCORE}_k,
%\nonumber\\
%\mathrm{SCORE}_{\text{HIGH}} \;=\; \sum_{k\in(3,4)}\mathrm{SCORE}_k.
%\label{eq:score_low_high}
%\end{align}

%Model configurations are compared using a Pareto analysis in the
%$(\mathrm{SCORE}_{\text{LOW}},\mathrm{SCORE}_{\text{HIGH}})$ space, with coverage displayed
%as an additional visual cue to distinguish robust from fragile trade-offs.

If either
$a$ or $b$ of a transition appears fewer than five times among the predicted risk values, the corresponding test is excluded to avoid evaluating the spline in regions supported by too few observations, which would make the estimate statistically unreliable. Note that the absence of a given risk level is not considered an error in itself, as no fully reliable ground-truth value exists to define the expected outcome at that level. Excluding such transitions also prevents attributing meaning to unsupported regions of the risk scale while preserving the validity of the evaluation on empirically observed risk values. Sensitivity to this threshold is assessed by repeating the evaluation with alternative minimum-count values.

\section{Risk Modeling Candidates}
\label{sec:candidates}

This section presents the different risk modeling approaches implemented in this article. This involves expert-based modeling, statistical modeling, predictive modeling, and generative modeling. All methods are designed to produce a risk level between 0 and 4 and are evaluated using the monotonic framework explained in Section~\ref{sec:mono}. The main characteristics, advantages, and disadvantages of each model are summarized in Table~\ref{tab:models_comparison}.

\begin{table*}[ht]
\centering
\caption{Comparison of Risk Modeling Candidates.}
\label{tab:models_comparison}
\begin{tabularx}{\textwidth}{lXX}
\toprule
Model & Advantages & Disadvantages \\
\midrule
DFE & - Operationally adopted (field use) & - No learning from data \newline - Ignores socio-economic data \newline - Requires expert calculation \newline - Requires regional adaptation \\ \hline
Poisson & - Interpretable baseline for counts \newline - Handles sparsity well & - Linear assumptions \newline - Limited capacity for complex patterns \\ \hline
Week-Max & - Easy to implement \newline - Captures seasonal peaks & - Poor at predicting highly stochastic events \newline - No daily weather sensitivity \\ \hline
Persistence & - Easy to implement \newline - Strong temporal correlation basis & - Poor at predicting highly stochastic events \newline - Reactive only (lag) \\ \hline
Logistic Regression & - Probabilistic classification \newline - Optimized for imbalance & - Limited feature interaction \newline - Struggles with ordinal constraints \\ \hline
GRU & - Captures temporal dependencies & - Requires training and architecture optimization \newline - Requires hyperparameter tuning \newline - Dependent on the given objective \\ \hline
FARS & - Contextual reasoning (Agents) \newline - Multi-perspective synthesis \newline - Explanability & - Risk of non-reproducibility \newline - High computational cost \newline - Dependent on predictive model performance \\
\bottomrule
\end{tabularx}
\end{table*}

\subsection{Expert Modeling}

To establish a fair baseline and enable a meaningful comparison between data-driven and expert-based modeling approaches, the DFE system presented was used.

\begin{definition}[DFE (Danger Final Expertisé)]\label{def:dfe}
	The DFE is an ordinal daily meteorological fire danger index ranging from 0 to 4, synthesized from weather and environmental indicators, and used operationally by experts during the summer fire season.
\end{definition}

This system offers greater relevance than less localized frameworks such as the FWI, as it is specifically adapted to our study region. Additional information about the DFE is available in~\cite{caron2026proofconceptmultitargetwildfire}.

\subsection{Statistical Modeling}

These models provide a comprehensive set of reference points, ranging from naive heuristics to established statistical learning methods tailored for imbalanced count and classification tasks. For each operational target, the supervised labels are obtained by discretizing the corresponding continuous variable using a quantile-based scheme with thresholds $[0.5, 0.75, 0.95]$, yielding an ordinal scale from 0 to 4. These include:

\begin{itemize}
	\item \textbf{Poisson Regression:} A generalized linear model fitted individually for each zone. It incorporates a sample weighting scheme to balance the representation of the zero class (no event) versus positive classes, with the optimal class-0 proportion tuned on a validation set.

	\item \textbf{Week-Max Heuristic:} A rule-based model that predicts the maximum historical value observed for a given zone and week of the year. This captures seasonal peaks and worst-case scenarios specific to each location and time of year.

	\item \textbf{Persistence:} A simple temporal baseline (implemented via the shift threshold mode) that assumes the future state will mirror the most recent past state, effectively shifting the target variable by one time step.

	\item \textbf{Logistic Regression:} A multiclass classification model trained with a similar class-balancing strategy as the Poisson model. It tunes the proportion of the majority class (class 0) to optimize the Intersection over Union (IoU) score, addressing the significant class imbalance inherent in the data.
\end{itemize}

\subsection{Predictive Modeling}

\subsubsection{Unique Predictive Modeling}
Each operational clustered target (number of fires, resources, and intervention time) is independently predicted using a stacked Gated Recurrent Unit (GRU) neural network, whose architecture is detailed in~\cite{caron2026proofconceptmultitargetwildfire}. This choice ensures consistency with the proof of concept and has the advantage of faster training compared to more complex models. The supervised targets are obtained by discretizing each corresponding continuous operational variable using a quantile-based scheme with thresholds $[0.5, 0.75, 0.95]$, yielding an ordinal scale from 0 to 4. The evaluation of more complex predictive models is discussed in Section~\ref{sec:Disc}. The models process sequential features over an 11-day window. The models are trained on zone-aggregated datasets, with class imbalance addressed by under-sampling class 0, recursively testing sampling proportions from 0.05 to 1.00. Model outputs are ordinal risk classes for each target for same-day forecasting. This means that, for each zone and each day, the models produce a predicted ordinal value for every target.

\subsubsection{Aggregative Predictive Modeling}
In addition to evaluating each target-specific GRU model separately, we define a linear aggregation baseline that combines the three predicted risk classes (fire, resources, and time) with the DFE risk level when available (horizon 0). The aggregated score is defined as a weighted sum in Equation~\ref{eq:aggregate_gru}.
\begin{multline}
	\label{eq:aggregate_gru}
	S_{\mathrm{lin}}
	= w_{\mathrm{fire}}\,\mathrm{GRU}_{\mathrm{fire}}
	+ w_{\mathrm{res}}\,\mathrm{GRU}_{\mathrm{res}}\\
	+ w_{\mathrm{time}}\,\mathrm{GRU}_{\mathrm{time}}
	+ w_{\mathrm{dfe}}\,\mathrm{DFE},
\end{multline}
where each weight belongs to the discrete set $\{0.0,\;0.25,\;0.5,\;0.75,\;1.0\}$. The weights are normalized such that their sum equals 1. The resulting aggregated score is then rounded to the nearest integer, ensuring it remains within the ordinal scale of 0 to 4.
This aggregated risk level is then evaluated using the monotonic framework described in Section~\ref{sec:mono}, by fitting Equation~\ref{eq:eval_lin}
\begin{equation}
	\label{eq:eval_lin}
	Y
	= f\!\left(S_{\mathrm{lin}}\right)
	+ Z
	+ D
	+ \varepsilon.
\end{equation}
Among all tested weight combinations, the best-performing configuration over the three targets is retained.

\subsection{Agentic Reasoning for Operational Risk Synthesis}

%This section introduces the agentic reasoning layer that enables the transformation of multiple weakly correlated predictive signals into a coherent operational risk decision.
The previously presented models remain limited as they each capture only a single facet of risk (e.g., number of fires, resource usage, or intervention time), while their aggregation is constrained by the assumption of linearity. To address these limitations, we introduce a multi-agent system designed to analyze the performance of each GRU model and aggregate its predicted risks into a single final value. The multi-agent structure is therefore designed to organize, constrain, and validate the reasoning process that leads from raw predictions to an operationally meaningful risk level.

\begin{figure*}[htbp]
	\centering
	\includegraphics[width=\linewidth]{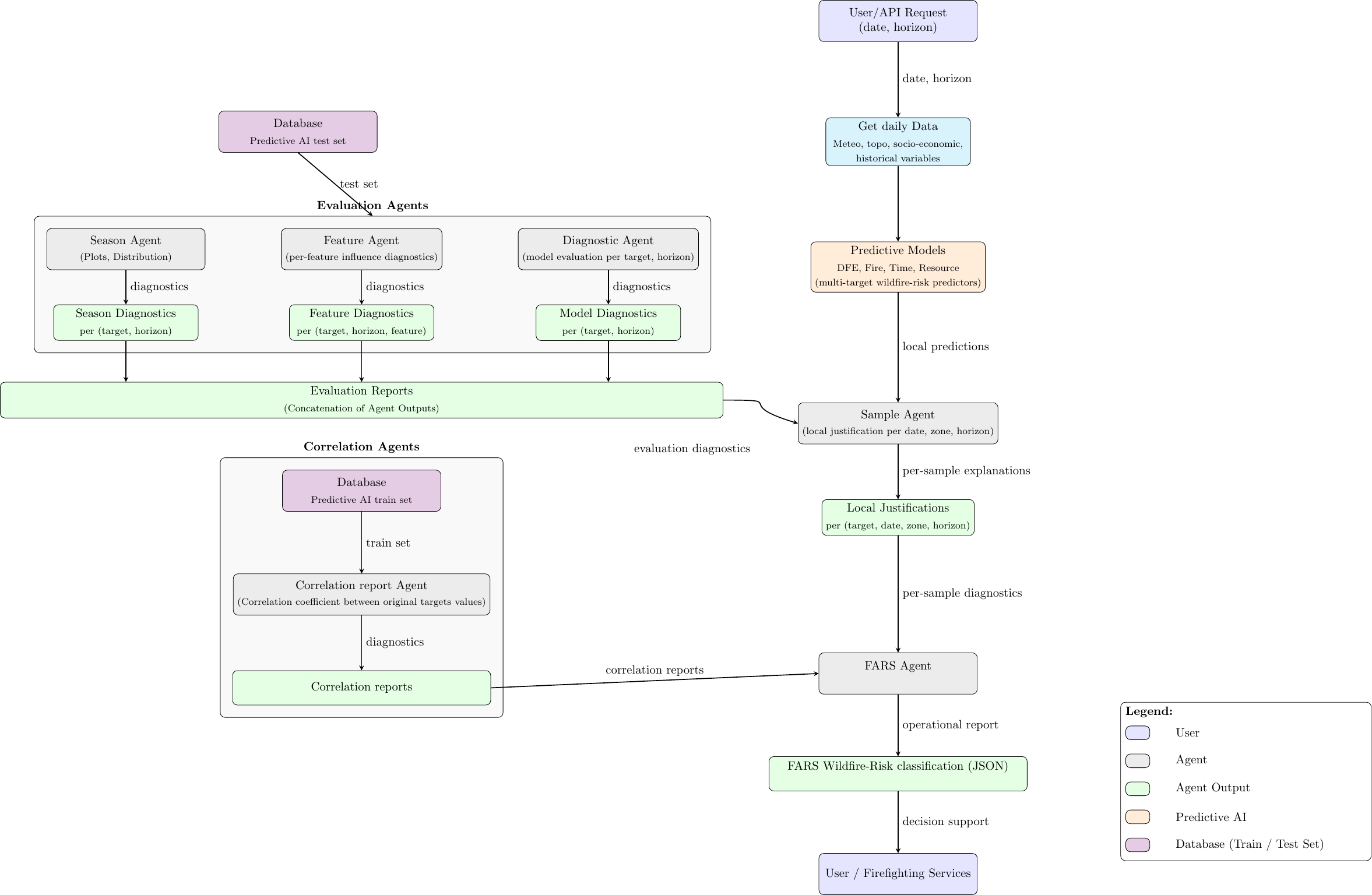}
	\caption{Overview of the generative multi-agent layer of the FARS.
		A user request for a specific date and horizon drives local justifications,
		and model diagnostics. Predictive models feed the evaluation agents, the sample agent gives a justification of the prediction, and the FARS Agent combines these elements to classify the current situation into 5 classes.}
	\label{fig:ag}
\end{figure*}

Figure~\ref{fig:ag} illustrates how these agents cooperate to ensure structural coherence, model reliability assessment, local justification, and contextual integration before producing the final wildfire-risk report. In this study, tests are performed at horizon 0 (i.e., producing a risk assessment for the current day). Accordingly, the multi-agent system, named \textit{FARS}, takes as input the operational predictions of each model as well as the official DFE value when it is available. For higher forecast horizons, the same architecture can incorporate a predicted DFE value (which yields a strong fit; see~\cite{caron2026proofconceptmultitargetwildfire}).

It is worth noting that, beyond aggregating the outputs of different models, LLM-based systems also provide capabilities for generating textual situation reports and recommendations. However, only the final aggregated risk value is considered in this study.

The FARS system includes six specialized agents. I) The \textbf{Correlation Agent} analyses historical data to characterize structural relationships between wildfire-risk targets (fires, intervention time, resources, DFE), identifying where dependencies are strong or weak across zones and providing the system with a reference map for coherent signal synthesis. II) The \textbf{Seasonality Agent} assesses whether predictive models capture the fundamental spatio-temporal structure of wildfire risk by identifying recurring temporal dynamics (risk peaks, seasonal transitions) and spatial heterogeneity. III) The \textbf{Diagnostic Agent} interprets standard evaluation metrics (macro F1-score, precision, recall, IoU) to determine model reliability across zones, identifying patterns such as class confusions and systematic biases. IV) The \textbf{Feature Agent} uses SHAP-based feature importance to reveal which input variables drive predictions, detecting inconsistencies or unexpected influences per zone and risk class. V) The \textbf{Sample Agent} explains each individual prediction by confronting it with diagnostics from the Feature, Seasonality, and Diagnostic agents, verifying coherence with local feature behaviour and spatio-temporal context. VI) The \textbf{FARS Agent} synthesizes all outputs to produce a single operational risk decision, structured into two components: one for risk classification and uncertainty estimation, and another for report generation. The result is an operational report that explains the situation, justifies the risk level, and provides recommendations for firefighting services.

\section{Evaluation}
\label{sec:eval}
\subsection{Experimental Setup}

We selected the period from 07 July to 25 September 2023, which is the maximal contiguous period where all operational targets and predictors are jointly available without missing data.

FARS uses the GPT-4.1 model from OpenAI (GPT-5.2 for the classifier), with a maximum context length of 4,000 tokens (and 20 tokens for the classifier) and a temperature parameter set to 1.0. GPT-4.1 was chosen for its reasoning stability across long diagnostic chains, while GPT-5.2 was selected for the final classification step due to its improved calibration on ordinal classification tasks. This temperature value was deliberately chosen to encourage diversity in the reasoning produced by diagnostic agents: lower temperatures tend to collapse outputs toward stereotypical, repetitive patterns, whereas a temperature of 1.0 allows the agents to explore a broader range of plausible interpretations when synthesizing heterogeneous signals into a coherent risk assessment. We acknowledge that training diagnostics on the same dataset as evaluation introduces a potential information leakage risk. However, we ensured that (1) diagnostic agents never access ground-truth values, (2) no date-specific information is used in agent prompts, and (3) the final classifier operates solely on aggregated diagnostic outputs.

\subsection{Results}

Table~\ref{tab:pareto} reports the monotonic evaluation performance (SCORE$_{k_1}$--SCORE$_{k_4}$) of each configuration for Fire, Resources, and Time, as well as the coverage ($k_1$--$k_4$) of risk transitions, indicating how well the predicted risk aligns with observed operational load and how much of the risk scale is effectively explored.

\begin{table*}[ht]
\centering
\scriptsize
\caption{MONOTONIC SCORE$_{k_1}$ – SCORE$_{k_4}$ AND COVERAGE BY CONFIGURATION AND TARGET (F=FIRE , R=RESOURCES , T=TIME ). COVERAGE SHOWN IN RIGHTMOST COLUMNS; TUPLE FORMAT (F, R, T) WHEN TARGET- DEPENDENT.}
\renewcommand{\arraystretch}{1.0}
\resizebox{\textwidth}{!}{%
\begin{tabular}{l|ccc|ccc|ccc|ccc|cccc}
\hline
\multirow{2}{*}{Config}
& \multicolumn{3}{c|}{SCORE$_{k_1}$}
& \multicolumn{3}{c|}{SCORE$_{k_2}$}
& \multicolumn{3}{c|}{SCORE$_{k_3}$}
& \multicolumn{3}{c|}{SCORE$_{k_4}$}
& \multicolumn{4}{c}{Coverage} \\
\cline{2-17}
& F & R & T & F & R & T & F & R & T & F & R & T & $k_1$ & $k_2$ & $k_3$ & $k_4$ \\
\hline

DFE
& 0.09 & 0.19 & -0.03
& 0.33 & 0.37 & 0.17
& 0.56 & 0.73 & 0.28
& 0.95 & 1.29 & 0.54
& 334 & 150 & 76 & 13 \\

\hline

GRU+DFE Agg.
& 0.04 & 0.10 & 0.07
& 0.09 & 0.21 & 0.17
& 0.53 & 0.73 & 0.26
& 1.78 & 1.77 & 0.43
& 267 & 132 & 41 & 5 \\

GRU (mono)
& 0.11 & 0.02 & 0.11
& 0.90 & 0.22 & 0.28
& -- & 0.70 & --
& -- & -- & --
& \footnotesize(133,179,123) & \footnotesize(19,86,40) & \footnotesize(0,6,0) & 0 \\

\hline
FARS
& -0.06 & 0.00 & -0.19
& 0.06 & 0.04 & -0.02
& 0.31 & 0.26 & -0.19
& 0.46 & 0.48 & 0.03
& 318 & 150 & 73 & 12 \\

FARS w/o F,T
& 0.12 & 0.23 & -0.06
& 0.28 & 0.37 & 0.11
& 0.63 & 0.99 & 0.15
& 1.02 & 1.00 & 0.36
& 336 & 136 & 71 & 11 \\

FARS w/o R
& 0.10 & 0.12 & 0.06
& 0.28 & 0.23 & 0.16
& 0.54 & 0.60 & 0.25
& 0.73 & 1.33 & 0.42
& 322 & 149 & 74 & 12 \\

FARS No Agents
& 0.06 & 0.01 & -0.09
& 0.25 & 0.08 & 0.09
& 0.53 & 0.41 & 0.08
& 0.86 & 1.02 & 0.23
& 334 & 150 & 77 & 12 \\

FARS w/o DFE
& 0.1 & 0.09 & -0.18
& 0.40 & 0.33 & -0.41
& 0.92 & 0.95 & -0.56
& -- & -- & -0.31
& 251 & 98 & 9 & 2 \\

\hline
Persistence
& -0.41 & -0.61 & -0.08
& -0.28 & -0.70 & -0.10
& -0.81 & -0.79 & -0.05
& -- & -- & 0.09
& \footnotesize(140,84,84) & \footnotesize(33,26,35) & \footnotesize(8,6,9) & \footnotesize(1,0,3) \\

MaxWeek
& -0.33 & -0.43 & -0.25
& -0.46 & -0.63 & -0.51
& -0.69 & -0.72 & -0.50
& -0.76 & -0.96 & 0.02
& \footnotesize(283,359,319) & \footnotesize(153,221,187) & \footnotesize(69,90,97) & \footnotesize(7,28,35) \\

Poisson
& 0.22 & -0.60 & -0.46
& 0.53 & -0.57 & -0.26
& -- & -0.64 & -0.12
& -- & -0.96 & -0.48
& \footnotesize(245,278,260) & \footnotesize(15,68,108) & \footnotesize(2,33,40) & \footnotesize(1,10,10) \\

LR
& 0.11 & -2.00 & -1.11
& -- & -3.58 & -2.38
& -- & -- & -2.78
& -- & -- & --
& \footnotesize(201,267,195) & \footnotesize(3,11,41) & \footnotesize(0,3,1) & \footnotesize(0,2,0) \\

\hline
\end{tabular}}
\label{tab:pareto}
\end{table*}

\paragraph{Coverage is necessary but not sufficient}
Several statistical and heuristic baselines such as \textit{MaxWeek},
\textit{Persistence}, \textit{Poisson}, and \textit{LR} exhibit very large
coverage of the risk scale. For instance,  \textit{MaxWeek} reaches coverage values of
$(283,359,319)$ in $k_1$ and $(153,221,187)$ in $k_2$ for
(Fire, Resources, Time), which is comparable or even superior to the DFE.
However, these configurations systematically produce strongly negative
SCORE values (e.g.,  \textit{MaxWeek}: $-0.33$, $-0.43$, $-0.25$ in $k_1$ and
$-0.46$, $-0.63$, $-0.51$ in $k_2$).
This demonstrates that exploring the full ordinal scale of risk does not guarantee that the signal behaves monotonically with respect to operational load. These models distribute classes well, but the ordering of these classes is not structurally aligned with increases in operational activity.

\paragraph{The DFE is the only naturally well-distributed and monotone signal}
The DFE achieves the most extensive and balanced coverage
(334, 150, 76, 13 transitions for $k_1$ to $k_4$) while maintaining
consistently positive SCORE values that increase with the transition order
(e.g., Fire: $0.09 \rightarrow 0.56$, Resources: $0.19 \rightarrow 0.73$).
It is the only system that simultaneously explores the whole risk scale and
maintains a coherent monotonic behavior across all targets.
However, its SCORE values remain moderate in high transition compared to the best GRU-based
aggregations, especially for Fire and Intervention Time in $k_4$, confirming that
DFE primarily captures environmental severity rather than the full complexity
of operational load. Additionally, DFE reaches a negative score for the Time target at $k_1$ ($-0.03$).

\textit{Resources} is not the most difficult
target to explain. On the contrary, it is the one for which the DFE
obtains some of its highest SCORE values
(e.g., $0.19$ in $k_1$, $0.37$ in $k_2$, $0.73$ in $k_3$ and $1.29$ in $k_4$).
This is not because resource deployment is easier to predict,
but because it is a decision-driven variable that is operationally guided
by the DFE itself.
In contrast, Fire occurrence and Intervention Time reflect more stochastic
and physical processes that are only partially captured by meteorological
danger indices.

\paragraph{Mono-target GRU models learn a strong but local monotonic structure}
The GRU models obtain high SCORE values, particularly for Fire and
Intervention Time (e.g., Fire: $0.11$ in $k_1$ and $0.90$ in $k_2$),
but their coverage collapses for higher-order transitions
($(0,6,0)$ in $k_3$ and $0$ in $k_4$).
This indicates that the models capture a meaningful monotonic relationship,
but fail to produce a risk signal that is well distributed across the ordinal
scale. They learn the relationship locally, but do not generate an
operationally exploitable risk scale.

\paragraph{Linear aggregation of GRU and DFE produces the strongest explanatory signal but with very low robustness}
The configuration \textit{All GRU\_F1.0-R0.0-T0.0\_DFE1.00}
achieves the highest SCORE values of the entire study
(e.g., $1.78$ and $1.77$ in $k_4$ for Fire and Resources).
However, the coverage in $k_4$ is only $5$ transitions.
This shows that when complementary signals are combined,
the explanatory power can be extremely high, but only on a very limited
portion of the risk scale. The signal is locally spectacular but
structurally incomplete.

\paragraph{FARS reveals a fundamental property of agentic aggregation}
The various FARS configurations provide particularly insightful results.
The full FARS configuration yields weak or negative SCORE values in lower
orders (e.g., $-0.06$, $0.00$, $-0.19$ in $k_1$),
while simplified variants such as \textit{FARS\_Without\_R} or
\textit{FARS\_Without\_F\_T} obtain much higher SCORE values
(up to $1.02$ in $k_4$).
Interestingly, \textit{FARS\_No\_Agents} sometimes performs better than the
complete agentic system.
This shows that the generative multi-agent reasoning does not correct
structural weaknesses of the predictive signal; it inherits them.
FARS acts as a revealer of the quality of upstream predictions rather than
a corrective layer.

\begin{figure}[ht]
	\centering
	\includegraphics[width=\columnwidth]{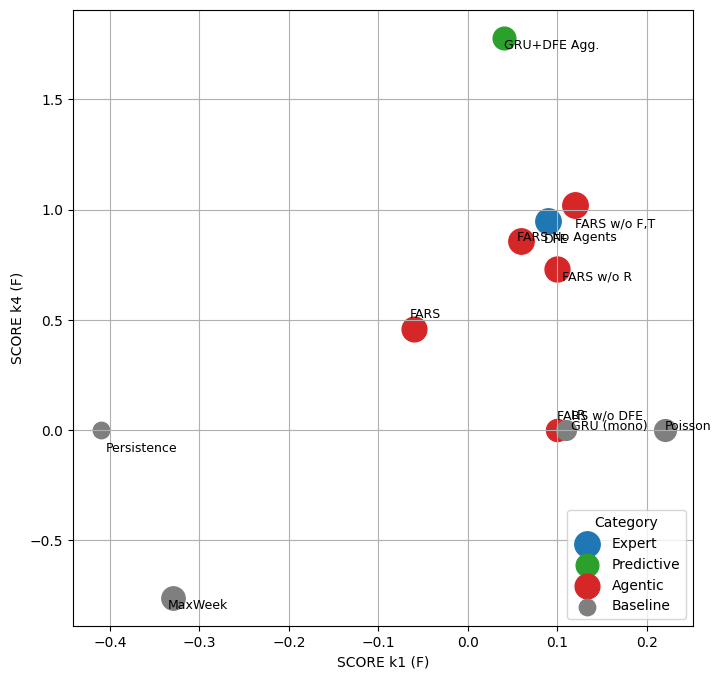}
	\caption{Pareto view of monotonic performance for the Fire target. Each point represents a configuration, with the x-axis showing SCORE$_{k_1}$ (monotonicity at low risk levels) and the y-axis showing SCORE$_{k_4}$ (monotonicity at high risk levels). Point size is proportional to coverage in $k_1$. Configurations in the upper-right quadrant exhibit good monotonic behavior across the full risk spectrum. The DFE, GRU+DFE, and the FARS w/o F,T aggregation show the best trade-offs. The joint analysis of $k_1$ (low-risk transitions) and $k_4$ (high-risk transitions) provides a clear visualization of the balance between sensitivity to minor fluctuations and robustness in extreme conditions.}
	\label{fig:pareto}
\end{figure}

\paragraph{A balance between low-order and high-order transitions}
Figure~\ref{fig:pareto} highlights a fundamental balance that a valid risk signal must satisfy: it must remain monotonic both for low-order transitions ($k_1$), corresponding to small increases in risk, and for high-order transitions ($k_4$), corresponding to extreme situations. Models that perform well only in $k_1$ capture local consistency but fail to structure the upper part of the ordinal scale, while models that score highly in $k_4$ often do so on a very limited portion of the risk spectrum. A reliable operational risk indicator must therefore jointly maintain monotonic coherence across weak and strong transitions while effectively covering the entire range of risk levels. This balance, visible in the upper-right region of the figure with large point sizes, characterizes signals whose ordinal structure meaningfully explains operational load across the full spectrum. The three most performant configurations are the DFE, the GRU+DFE aggregation, and the FARS variant without Fire and Time inputs, as they occupy the upper-right region of the plot while maintaining relatively large point sizes. This FARS configuration shows the best balance between the two types of transitions. This observation suggests that aggregating heterogeneous types of risk signals may be a promising direction for future work, as combining complementary sources of information appears to improve the balance of monotonic behavior across the full range of risk transitions.

\paragraph{Analysis of classical AI metrics.}

Table~\ref{tab:metrics} presents the classical predictive metrics, obtained on the clustered target Fire for all configurations. These metrics are usually used in wildfire evaluation~\cite{ai6100253}.

\begin{table*}[ht]
	\caption{Predictive metrics for the target Fire across configurations.}
	\centering
	\small
	\begin{tabular}{lcccccc}
		\hline
		Config                 & MAE    & IoU    & F1\_Macro & F1\_Binary & Rec & Prec \\
		\hline
		GRU              & 0.3139 & 0.2644 & 0.3537    & 0.2114         & 0.1250  & 0.6842   \\
		DFE                    & 1.3553 & 0.0724 & 0.1314    & 0.3675         & 0.6731  & 0.2527   \\
		FARS\_w/o\_F\_T    & 1.3647 & 0.0577 & 0.1045    & 0.3598         & 0.6538  & 0.2482   \\
		FARS\_No\_Agents       & 1.3289 & 0.0590 & 0.1060    & 0.3836         & 0.6731  & 0.2682   \\
		FARS\_w/o\_DFE     & 0.6049 & 0.1505 & 0.2168    & 0.4257         & 0.4135  & 0.4388   \\
		FARS                   & 1.2895 & 0.0615 & 0.1099    & 0.3652         & 0.6058  & 0.2614   \\
		FARS\_w/o\_R       & 1.2981 & 0.0615 & 0.1105    & 0.3818         & 0.6442  & 0.2713   \\
		LR     & 0.5602 & 0.1418 & 0.2150    & 0.0561         & 0.0288  & 1.0000   \\
		Persistence            & 0.4388 & 0.1622 & 0.2127    & 0.1168         & 0.0769  & 0.2424   \\
		MaxWeek                & 1.2744 & 0.0603 & 0.1090    & 0.4185         & 0.6538  & 0.3077   \\
		Poisson                & 0.5038 & 0.1484 & 0.2166    & 0.0672         & 0.0385  & 0.2667   \\
		\hline
	\end{tabular}
	\label{tab:metrics}
\end{table*}

This table highlights a fundamental mismatch between the nature of a risk
signal and the nature of the supervised target used for evaluation.
The DFE, which is designed as a continuous ordinal indicator of
environmental danger, obtains very low predictive scores when directly
compared to the discrete target Fire
(e.g., MAE $=1.355$, IoU $=0.072$, F1\_Macro $=0.131$).
This does not indicate that the DFE is a poor risk signal, but rather that
it is not intended to predict the exact number of fire events.

Machine learning models, such as the GRU-based model, are explicitly trained
to fit this discrete clustered target and therefore obtain much better
predictive metrics (e.g., MAE $=0.314$, IoU $=0.264$, F1\_Macro $=0.354$).
These models are structurally advantaged in this evaluation because they are
optimized to reproduce the target variable, whereas the DFE is not.

This comparison illustrates why classical predictive metrics are not appropriate for assessing the quality of a risk signal: they reward the ability to fit a discrete stochastic outcome rather than the ability to produce a meaningful continuous indicator of operational danger.

\paragraph{The issue is not prediction, but signal shape}
The best score values do not come from the most accurate predictive models, but from the configurations that produce the best ordinal distribution of
risk levels. This represents a conceptual shift: a good risk model is not one that best
fits the target, but one whose ordinal scale consistently explains increases
in operational load across the entire risk spectrum.

\section{Discussion}\label{sec:Disc}

The proposed monotonic evaluation framework demonstrates that the DFE, one of the worst predictors according to classical metrics, behaves as one of the most coherent operational risk signals under monotonic evaluation. This emphasizes a fundamental point: in operational load prediction, the objective is not to fit a highly stochastic target such as fire counts, but to produce a signal whose ordinal structure maintains a monotonic relationship with true operational outcomes. This perspective fundamentally changes the optimization paradigm for predictive models in this domain.

However, a critical limitation must be acknowledged: the DFE actively \emph{influences} resource deployment decisions. When the DFE indicates a higher danger, operational protocols require increased standby resources and pre-positioned units. The observed correlation between DFE levels and deployed resources is therefore partly \emph{circular}: the DFE prescribes rather than explains resource needs. This does not invalidate its operational utility but complicates interpretation, as a portion of the alignment reflects that resource allocation is directly conditioned on the DFE itself. This self-fulfilling property is less pronounced for Fire and Time targets, which depend on stochastic environmental factors beyond the control of operational protocols.

The aggregation of complementary signals produces strong results, especially when predictive outputs are combined with the DFE\@. However, the full FARS system inherits the structural weaknesses of its upstream predictive signals rather than correcting them. The results show clearly that, in its current version, agentic aggregation cannot compensate for fundamental limitations in the quality of individual predictions. This effect becomes even more evident when observing the increase in scores obtained after removing certain targets, particularly Fire and Time. This suggests that the way heterogeneous risk components are combined directly affects the structural quality of the resulting risk signal. Future work should therefore focus on improving predictive signal quality before attempting higher-level generative aggregation. Furthermore, the risk score calculation presented in Section~\ref{sec:mono} will be integrated into the Diagnostic Agent to improve its ability to link predicted risk levels with operational reality. Finally, contextual agents capable of interpreting weather alerts, event calendars, and local operational constraints will be added to the system.

The experiments are localized to the Alpes-Maritimes department during summer 2023; no claim of generalization is made. Nevertheless, the framework can be generalized to other ordinal risk systems where the goal is to produce a score that reflects an increase in an objective operational quantity. An immediate extension of the framework would be to impose additional structural constraints on the ordinal scale, such as requirements on the lowest risk level and minimum expected increases between successive levels. The score is designed to summarize transitions along the ordinal risk scale, while a transition matrix could provide a compact way to visualize pairwise effects, local monotonicity violations, and effective coverage. A further important extension would be to normalize transition deltas against an ideal reference ordinal scheme, so that scores become more demanding but also more interpretable relative to a near-perfect system. Code on~\href{https://github.com/NicolasCaronPro/Risk-Is-Not-the-Target-A-Monotonic-Framework-for-Evaluating-Wildfire-Operational-Risk-Signals}{github} 

From a methodological standpoint, the use of LLM-based agents introduces an inherent source of stochasticity in the risk assessment process. While we fixed the temperature parameter and seed where possible, the non-deterministic nature of large language models means that repeated runs may produce slightly different outputs. This variability is acceptable for diagnostic and explanatory purposes, but the final classification step relies on a more calibrated model to ensure reproducibility.

\section{Conclusion}
\label{sec:concl}

The generation of AI-based operational wildfire risk signals remains largely unexplored in the literature, especially regarding how such systems can be evaluated without bias. To the best of our knowledge, we are the first to introduce a monotonic evaluation framework that assesses risk signals based on their coherence with observed operational load rather than their ability to predict discrete events. Through comparison of expert indices, statistical models, predictive AI, and the FARS multi-agent system, we show that the key property of a risk system is the structure of its ordinal scale and its monotonic relationship with operational outcomes. The results highlight both the potential and current limitations of hybrid predictive–generative approaches, and point toward future work on improving predictive signal quality, contextual agent integration, and real-time deployment within operational firefighting workflows. Based on this work, we can consider that a good risk model does not predict fires accurately, but one whose ordinal scale meaningfully explains operational dynamics. This work opens the door to a new way of constructing reliable and robust risk AI models.

\bibliographystyle{IEEEtran}  % Style classique numerot
\bibliography{main}

\end{document}